\documentclass{article}

\usepackage{amsmath,amsfonts,amssymb,times,graphicx,natbib,algorithm,algorithmic,hyperref}

\usepackage[accepted]{whi2016}

\icmltitlerunning{Interpretable Two-level Boolean Rule Learning for Classification}

\DeclareMathAlphabet{\mathbbb}{U}{bbold}{m}{n}
\DeclareMathOperator*{\argmin}{arg\,min}

\newcommand{\ANDL}{\bigwedge}
\newcommand{\ORL}{\bigvee}
\newcommand{\uth}{^{\rm{th}}}
\newcommand{\kron}{\mathbbb{1}}
\newcommand{\DC}{{\rm{DC}}}

\begin{document}

\twocolumn[
\icmltitle{Interpretable Two-Level Boolean Rule Learning for Classification}

\icmlauthor{Guolong Su}{guolong@mit.edu}
\icmladdress{Massachusetts Institute of Technology,
            50 Vassar St., Cambridge, MA 02139 USA}
\icmlauthor{Dennis Wei}{dwei@us.ibm.com}
\icmlauthor{Kush R. Varshney}{krvarshn@us.ibm.com}
\icmlauthor{Dmitry M. Malioutov}{dmalioutov@us.ibm.com}
\icmladdress{IBM Thomas J. Watson Research Center,
            1101 Kitchawan Rd., Yorktown Heights, NY 10598 USA}

\icmlkeywords{Interpretable Classifier, Linear Programming Relaxation}

\vskip 0.3in
]

\begin{abstract}
As a contribution to interpretable machine learning research, we develop a novel optimization framework for learning accurate and sparse two-level Boolean rules. We consider rules in both conjunctive normal form (AND-of-ORs) and disjunctive normal form (OR-of-ANDs). A principled objective function is proposed to trade classification accuracy and interpretability, where we use Hamming loss to characterize accuracy and sparsity to characterize interpretability. We propose efficient procedures to optimize these objectives based on linear programming (LP) relaxation, block coordinate descent, and alternating minimization. Experiments show that our new algorithms provide very good tradeoffs between accuracy and interpretability.
\end{abstract}

\section{Introduction}\label{sec:Introduction}

In applications where machine learning is used to aid human decision-making, it is recognized that interpretability of models is an important objective for establishing trust, adoption and safety and for offering the possibility of auditing and debugging \citep{freitas2014comprehensible,varshney2016safety,lima2009domain,letham2012building,vellido2012making}.  Interpretable models can be learned directly from data or can result from the approximation of black-box models \cite{craven1996extracting,baesens2003using}.

Boolean rules are considered to be one of the most interpretable classification models \cite{freitas2014comprehensible}. A one-level rule is a conjunctive clause (AND-rule) or disjunctive clause (OR-rule) whereas a two-level rule is a conjunctive normal form (CNF; AND-of-ORs) or disjunctive normal form (DNF; OR-of-ANDs).  A mathematical proxy for interpretability of Boolean rules is \emph{sparsity}, i.e.\ the total number of features used in the rule \cite{feldman2000minimization}.

Learning accurate and sparse two-level Boolean rules is a considerable challenge since it is combinatorial \cite{kearns1987learnability}. Even the simpler problem of learning a one-level rule is NP-hard \cite{malioutov2013exact}. Unlike one-level rules, two-level rules can represent any Boolean function of the input features \cite{furnkranz2012foundations}; this expressiveness of two-level rules also suggests that they are more challenging to learn than one-level rules. Due to this complexity, most existing solutions focus on heuristic and greedy approaches.

The main contribution of this paper is to introduce a unified optimization framework for learning two-level Boolean rules that achieve good balance between accuracy and interpretability, as measured by the sparsity of the rule.  The objective function is a weighted combination of (a) classification errors quantified by Hamming distance between the current rule and the closest rule that correctly classifies a sample and (b) sparsity. Based on this formulation, block coordinate descent and alternating minimization algorithms are developed, both of which use an LP relaxation approach. Experiments show that two-level rules can have considerably higher accuracy than one-level rules and may show improvements over cutting edge approaches.

The two-level Boolean rules in this work are examples of sparse decision rule lists \cite{rivest1987learning}, which have been extensively studied in various fields.  A number of strategies have been proposed \cite{furnkranz2012foundations}: covering \cite{rivest1987learning,clark1989cn2,cohen1995fast,furnkranz1999separate}, bottom-up \cite{salzberg1991nearest,domingos1996unifying}, multi-phase \cite{liu1998integrating,knobbe2008local}, and the distillation of trained decision trees into decision lists \cite{quinlan1987simplifying}. Unlike our proposed approach, the above strategies lack a single, principled objective function to drive rule learning. Moreover, they employ heuristics that leave room for improvements on both accuracy and rule simplicity.  In addition to the symbolic approaches above, Bayesian approaches in \citet{letham2012building} and \citet{WangRuWaBigData14} apply approximate inference algorithms to produce posterior distributions over decision lists; however, the assignment of prior and likelihood may not always be clear, and certain approximate inference algorithms may have high computational cost.

There has been some prior work on optimization-based formulations for rule learning, the most relevant being \citet{malioutov2013exact}, where an LP framework is proposed to learn one-level rules from which set covering and boosting are used to construct two-level rules. Although we apply this clause learning method as a component in our algorithms, our work has significant differences from \citet{malioutov2013exact}. As discussed earlier, two-level rules are significantly more expressive and much more challenging to learn than a one-level rule. In addition, the greedy style of the set covering method leaves room for improvement and the weighted combination of clauses in boosted classifiers reduces interpretability. Another work on DNF learning \cite{wang2015learning} provides a mixed integer program (MIP) formulation named OOA and a different heuristic formulation OOAx. The MIP in OOA is similar to our formulation with a different cost ($0$-$1$ error) but lacks an LP relaxation. OOAx is similar to the heuristic multi-phase strategy above.

\section{Problem Formulation}\label{sec:ProbFormulation}
We consider supervised binary classification given a training dataset of $n$ samples, where each sample has a label $y_i\in\{0,1\}$ and $d$ binary features{\footnote{We assume the negation of each feature is included as another input feature.}} $a_{i,j}\in\{0,1\}$ ($1{\le}j{\le}d$). The goal is to learn a classifier $\hat{y}(\cdot)$ in CNF (AND-of-ORs) that can generalize well from the training dataset.\footnote{The presentation focuses on CNF rules, but the proposed algorithms apply equally to learning DNF rules using De Morgan's laws.} In the lower level of the rule, each of $R$ clauses is formed by the disjunction of a selected subset of input features; in the upper level, the predictor is obtained as the conjunction of all clauses. Letting the decision variables $w_{j,r}\in\{0,1\}$ represent whether to include the $j\uth$ feature in the $r\uth$ clause, the clause and predictor outputs are given by
\begin{align}
\hat{v}_{i,r}&=\ORL_{j=1}^d \left(a_{i,j}w_{j,r}\right),\ {\rm{for}}\ 1\le i\le n,\ 1\le r\le R.\label{def-hatv}\\
\hat{y}_i&=\ANDL_{r=1}^{R}\hat{v}_{i,r},\ {\rm for}\ 1\le i\le n.\label{def-haty-hatv}
\end{align}

To mitigate the need for careful specification of the number of clauses $R$, we allow clauses to be ``disabled'' by padding the input feature matrix with a trivial ``always true'' feature $a_{i,0}=1$ for all $i$, with corresponding decision variables $w_{0,r}$ for all clauses.  If $w_{0,r}=1$, then the $r\uth$ clause has output $1$ and thus drops out of the upper-level conjunction in a CNF rule.  The sparsity cost for $w_{0,r}$, i.e.\ for disabling a clause, can be lower than other variables or even zero.

In learning Boolean rules, it is desirable to use a finer-grained measure of accuracy than the usual $0$-$1$ loss to distinguish between degrees of incorrectness and indicate where corrections are needed. Herein we propose measuring the accuracy of a rule on a single sample in terms of the minimal Hamming distance from the rule to an \emph{ideal} rule, i.e.\ one that correctly classifies the sample. The Hamming distance between two CNF rules is the number of $w_{j,r}$ that are different in the two rules. Thus the minimal Hamming distance represents the smallest number of modifications (i.e.\ negations) needed to correct a misclassification. 

For mathematical formulation, we introduce \emph{ideal clause outputs} $v_{i,r}$ with $1\le i\le n$ and $1\le r\le R$ to represent a CNF rule that correctly classifies the $i\uth$ sample. The values of $v_{i,r}$ are always consistent with the ground truth labels, i.e.\ $y_i=\ANDL_{r=1}^R v_{i,r}$ for all $1\le i\le n$. We let $v_{i,r}$ have a ternary alphabet $\{0,1,\DC\}$, where $v_{i,r}=\DC$ means that we ``don't care'' about the value of $v_{i,r}$. With this setup, if $y_i=1$, then $v_{i,r}=1$ for all $1\le r\le R$; if $y_i=0$, then $v_{i,r_0}=0$ for at least one value of $r_0$, and we can have $v_{i,r}=\DC$ for all $r\ne r_0$. In implementation, $v_{i,r}=\DC$ implies the removal of the $i\uth$ sample in the training or updating for the $r\uth$ clause, which leads to a different training subset for each clause.

For a given $v_{i,r}$, the minimal Hamming distance between the $r\uth$ clauses only of a CNF rule and an ideal rule can be derived as follows.  If $v_{i,r} = 1$, at most one positive feature needs to be included to produce the desired output, so the minimal Hamming distance is given by $\max\bigl\{0, 1-\sum_{j=1}^{d} a_{i,j} w_{j,r} \bigr\}$.  If $v_{i,r} = 0$, any $w_{j,r}$ with $a_{i,j} w_{j,r} = 1$ needs to be negated to be correct, resulting in a minimal Hamming distance of $\sum_{j=1}^{d} a_{i,j} w_{j,r}$.  Summing over $i, r$ and defining the sparsity cost as the sum of the numbers of features used in each clause, the problem is formulated as
\begin{align}
\min_{w_{j,r},\ v_{i,r}}\ & \sum_{i=1}^n\sum_{r=1}^R \Bigg[\kron_{v_{i,r}=1}\cdot\max\Bigg\{0, \Bigg(1-\sum_{j=1}^{d}a_{i,j}w_{j,r}\Bigg)\Bigg\}\nonumber\\
& + \kron_{v_{i,r}=0}\cdot \sum_{j=1}^{d}a_{i,j}w_{j,r}\Bigg]  + \theta\cdot\sum_{r=1}^R \sum_{j=1}^d w_{j,r}\label{HammingFormulationBothVW}\\
{\rm s.t.}\ &\ANDL_{r=1}^R v_{i,r}=y_i,\ \forall i,\label{HammingFormulationBothVWIdealV}\\
&\ v_{i,r}\in\{0,1,\DC\},\ w_{j,r}\in\{0,1\},\ \forall i,j,r.\nonumber
\end{align}
The ideal clause output constraint \eqref{HammingFormulationBothVWIdealV} requires that $v_{i,r} = 1$ for all $r$ if $y_i = 1$, as noted above.  For $y_i = 0$, $v_{i,r}=0$ needs to hold for at least one value of $r$ while all other $v_{i,r}$ can be $\DC$.  The Hamming distance is minimized by setting
\begin{equation}\label{def-r0}
v_{i,r_0}=0,\ {\rm{where}}\ \ r_0=\argmin_{1\le r\le R}\left(\sum_{j=1}^d a_{i,j} w_{j,r}\right).
\end{equation}

The binary variables $w_{j,r}$ can be further relaxed to $0\le w_{j,r}\le1$. However, the resulting continuous relaxation is generally non-convex for $R > 1$.  Additional simplifications are proposed in Section \ref{sec:Approach} to make the continuous relaxations more efficiently solvable.

Lastly, it can be seen that letting $R=1$ in formulation (\ref{HammingFormulationBothVW}) recovers the formulation for one-level rule learning in \citet{malioutov2013exact}.

\section{Optimization Approaches}\label{sec:Approach}

This section proposes a block coordinate descent algorithm and an alternating minimization algorithm to solve the regularized Hamming loss minimization in \eqref{HammingFormulationBothVW}.

\subsection{Block Coordinate Descent Algorithm}\label{subsec:BlockCoDes}
This algorithm considers the decision variables in a single clause ($w_{j,r}$ with a fixed $r$) as a block of coordinates, and performs block coordinate descent to minimize the Hamming distance objective function in (\ref{HammingFormulationBothVW}). Each iteration updates a single clause with all the other $(R-1)$ clauses fixed, using the one-level rule learning algorithm in \citet{malioutov2013exact}. We denote $r_0$ as the clause to be updated.

The optimization of (\ref{HammingFormulationBothVW}) even with $(R-1)$ clauses fixed still involves a joint minimization over $w_{j,r_0}$ and the ideal clause outputs $v_{i,r}$ for $y_i=0$ ($v_{i,r}=1$ for $y_i=1$ are 
fixed), so the exact solution could still be challenging. To simplify, we fix the values of $v_{i,r}$ for $y_i=0$ and $r\ne r_0$ to the actual clause outputs $\hat{v}_{i,r}$ in (\ref{def-hatv}) with the current $w_{j,r}$ ($r\ne r_0$). Now we assign $v_{i,r_0}$ for $y_i=0$: if there exists $v_{i,r}=\hat{v}_{i,r}=0$ with $r\ne r_0$, then this sample is guaranteed to be correctly classified and we can assign $v_{i,r_0}=\DC$ to minimize the objective in (\ref{HammingFormulationBothVW}); in contrast, if $\hat{v}_{i,r}=1$ holds for all $r\ne r_0$, then the constraint (\ref{HammingFormulationBothVWIdealV}) requires $v_{i,r_0}=0$.

This derivation leads to the updating process as follows. To update the $r_0\uth$ clause, we remove all samples that have label $y_i=0$ and are already predicted as $0$ by at least one of the other $(R-1)$ clauses, and then update the $r_0\uth$ clause with the remaining samples using the one-level rule learning algorithm.

There are different choices of which clause to update in an iteration. For example, we can update clauses cyclically or randomly, or we can try the update for each clause and then greedily choose the one with the minimum cost. The greedy update is used in our experiments.

The initialization of $w_{j,r}$ in this algorithm also has different choices. For example, one option is the set covering method, as is used in our experiments. Random or all-zero initialization can also be used.

\subsection{Alternating Minimization Algorithm}\label{subsec:AlterMin}
This algorithm alternately minimizes with respect to the decision variables $w_{j,r}$ and the ideal clause outputs $v_{i,r}$ in \eqref{HammingFormulationBothVW}. Each iteration has two steps: update $v_{i,r}$ with the current $w_{j,r}$, and update $w_{j,r}$ with the new $v_{i,r}$. The latter step is simpler and will be first discussed.

With fixed values of $v_{i,r}$, the minimization over $w_{j,r}$ is relatively straight-forward: the objective in (\ref{HammingFormulationBothVW}) is separated into $R$ terms, each of which depends only on a single clause $w_{j,r}$ with a fixed $r$. Thus, all clauses are decoupled in the minimization over $w_{j,r}$, and the problem becomes parallel learning of $R$ one-level clauses. Explicitly, the update of the $r\uth$ clause removes samples with $v_{i,r}=\DC$ and then uses the one-level rule learning algorithm.

The update over $v_{i,r}$ with fixed $w_{j,r}$ follows the discussion in Section \ref{sec:ProbFormulation}: for positive samples $y_i=1$, $v_{i,r}=1$, and for the negative samples $y_i=0$, $v_{i,r_0}=0$ for $r_0$ defined in (\ref{def-r0}) and $v_{i,r}=\DC$ for $r\ne r_0$. For negative samples with a ``tie'', i.e.\ non-unique $r_0$ in (\ref{def-r0}), tie breaking is achieved by a ``clustering'' approach. First, for each clause $1\le r_0\le R$, we compute its cluster center in the feature space by taking the average of $a_{i,j}$ (for each $j$) over samples $i$ for which $r_0$ is minimal in (\ref{def-r0}) (including ties). Then, each sample with a tie is assigned to the clause with the closest cluster center in $\ell_1$-norm among all minimal $r_0$ in (\ref{def-r0}).

Similar to the block coordinate descent algorithm, various options exist for initializing $w_{j,r}$ in this algorithm. The set covering approach is used in our experiments.

\section{Numerical Evaluation}\label{sec:Numerical}

\begin{table*}
\caption{10-fold Average Test Error Rates (unit: $\%$). Standard Error of the Mean is Shown in Parentheses.}\label{table:testrate}
\vskip 0.05in
\begin{center}
\begin{small}
\begin{sc}
\begin{tabular}{|c|c|c|c|c|c|c|c|c|}
\hline
Dataset          &  BCD          &   AM          &      OCRL    &  SC           &   DList       &   C5.0        &  CART         &  RIPPER\\\hline
ILPD        & $28.6(0.2)$   & $28.6(0.2)$   & $28.6(0.2)$  &$28.6(0.2)$    & $36.5(1.4)$   & $30.5(2.0)$   &  $32.8(1.3)$  &  $31.9(1.0)$\\\hline
Ionos        & $9.4(1.1)$    & $11.4(1.1)$   & $9.7(1.5)$   &$10.5(1.3)$    & $19.9(2.3)$   & $7.4(2.1)$    &  $10.8(1.2)$  &  $10.0(1.5)$\\\hline
Liver       & $37.1(3.2)$  & $39.1(2.5)$   & $45.8(2.2)$  &$41.7(2.8)$    & $45.2(2.6)$   & $36.5(2.4)$   &  $37.1(2.5)$   &  $35.4(2.2)$\\\hline
MAGIC             & $17.4(0.2)$  & $17.0(0.1)$   & $23.6(0.3)$  &$19.7(0.2)$    & $18.5(0.4)$   & $14.1(0.2)$   &  $17.8(0.3)$   &  $15.1(0.3)$\\\hline
Musk               & $7.5(0.3)$   & $3.6(0.6)$    & $8.4(0.1)$  & $8.1(0.3)$    & $13.6(0.5)$   & $3.1(0.4)$   &  $3.3(0.3)$   &  $3.7(0.2)$\\\hline
Parkin      & $12.8(2.2)$   & $15.9(2.9)$   & $16.4(2.1)$  &$14.9(1.9)$    & $25.1(3.3)$  & $16.4(2.7)$   &  $13.9(2.9)$   &  $10.7(1.8)$\\\hline
Pima        & $26.8(1.7)$   & $23.8(2.0)$   & $27.2(1.5)$  &$27.9(1.5)$    & $31.4(1.6)$   & $24.9(1.7)$   &  $27.3(1.5)$  &  $24.9(1.1)$\\\hline
Sonar     & $29.8(3.0)$   & $25.5(2.4)$   & $34.6(2.7)$  &$28.8(2.9)$    & $38.5(3.6)$  & $25.0(4.2)$  &  $31.7(3.5)$     &  $25.5(3.1)$\\\hline
Trans      & $23.8(0.1)$   & $23.8(0.1)$   & $23.8(0.1)$  &$23.8(0.1)$    & $35.4(2.4)$   & $21.7(1.2)$   &  $25.4(1.7)$  &  $21.5(0.8)$\\\hline
WDBC        & $6.2(1.2)$    & $6.5(0.9)$    & $9.3(2.0)$   &$8.8(2.0)$     & $9.7(0.8)$    & $6.5(1.1)$    &  $8.4(1.0)$   &  $7.4(1.2)$\\\hline
\hline
Ranking &$3.1$  &$3.2$  &$5.5$  &$4.9$  &$7.7$  &$2.6$  &$5.0$  &$2.8$  \\\hline
\end{tabular}
\end{sc}
\end{small}
\end{center}
\vskip -0.3in
\end{table*}

This section evaluates the algorithms with UCI repository datasets \cite{Lichman:2013}. To facilitate comparison with the most relevant prior work \cite{malioutov2013exact}, we use all $8$ datasets in that work. Each continuous valued feature is converted to binary using $10$ quantile thresholds. In addition, we use $2$ large datasets: MAGIC gamma telescope (MAGIC) and Musk version 2 (Musk).

The goal is to learn a DNF rule (OR-of-ANDs) from each dataset. We use stratified $10$-fold cross validation and then average the error rates. All LPs are solved by CPLEX version 12 \cite{IBMILOG}. The sparsity parameter $\theta$ is tuned between $10^{-4}$ and $50$ using a second cross validation within the training partition. 

Algorithms in comparison and their abbreviations are: block coordinate descent (BCD), alternating minimization (AM), one-level conjunctive rule learning (OCRL, equivalent to setting $R=1$ for BCD or AM) and set covering (SC), the last two from \citet{malioutov2013exact}, decision list in IBM SPSS (DList), decision trees (C5.0: C5.0 with rule set option in IBM SPSS, CART: classification and regression trees algorithm in Matlab's classregtree function), and RIPPER from \citet{cohen1995fast}. We set the maximum number of clauses $R=5$ for the BCD, AM, and SC algorithms, and set the maximum number of iterations in BCD and AM as $100$.

We first show the test error rates and the sparsity of the rules. The mean test error rates and the standard error of the mean are listed in Table \ref{table:testrate}. Due to space constraints, we refer the reader to \citet{malioutov2013exact} for results from other classifiers that are not interpretable; the accuracy of our algorithms is generally quite competitive with them. Table \ref{table:features} provides the $10$-fold average of the sparsity of the learned rules as a measure for interpretability. No features are counted if a clause is disabled. The last rows in these tables show the averaged ranking of each algorithm on each dataset.

\begin{table}
\caption{10-fold Average Numbers of Features}\label{table:features}
\vskip 0.05in
\begin{center}
\begin{small}
\begin{sc}
\setlength{\tabcolsep}{2pt}
\begin{tabular}{|c|c|c|c|c|c|c|}
\hline
Dataset            &   BCD         &   AM          &       SC       &   DList       &   C5.0    &  RIPPER     \\\hline
ILPD              & $0.0$         & $0.0$         &    $0.0$       &  $5.4$        & $45.5$    &  $7.0$       \\\hline
Ionos            &$12.4$         &$12.9$         &   $11.1$       &  $7.7$        & $13.6$    &  $6.0$       \\\hline
Liver             & $9.3$         & $7.7$         &    $5.2$       &  $2.2$        & $46.6$    &  $4.0$       \\\hline
MAGIC            & $11.4$        & $22.3$        &    $2.0$       &  $14.7$        & $366.7$    &  $110.0$       \\\hline
Musk               & $26.5$        & $63.7$        &    $18.3$      &  $15.9$        & $155.1$    &  $92.0$       \\\hline
Parkin           & $8.2$         &$12.6$         &    $3.2$       &  $2.1$        & $16.6$    &  $6.0$       \\\hline
Pima             & $2.2$         & $2.0$         &    $2.4$       &  $8.6$        & $38.2$    &  $5.0$       \\\hline
Sonar            &$14.2$         &$23.6$         &    $9.0$       &  $1.9$        & $27.3$    &  $8.0$       \\\hline
Trans           & $0.0$         & $0.0$         &    $0.0$       &  $3.8$        &  $6.7$    &  $5.0$       \\\hline
WDBC           &$13.6$         &$11.9$         &    $8.7$       &  $4.0$        & $15.8$    &  $6.0$       \\\hline
\hline
Ranking       &$3.1$      &$3.4$            &$2.2$         &$2.3$             &$6.0$        &$3.4$ \\\hline
\end{tabular}
\end{sc}
\end{small}
\end{center}
\vskip -0.3in
\end{table}

Table \ref{table:testrate} shows that two-level rules obtained by our algorithms (BCD and AM) are more accurate than the one-level rules from OCRL for almost all datasets, which demonstrates the expressiveness of two-level rules. Among optimization-based two-level rule learning approaches, BCD and AM generally have superior accuracy to SC. All these approaches substantially outperform DList in terms of accuracy on all datasets. Compared with C5.0, BCD and AM obtain rules with much higher interpretability (many fewer features) and quite competitive accuracy. Compared with CART, BCD has higher or equal accuracy on all datasets except for Musk, and AM is also superior overall. RIPPER appears to be slightly stronger than BCD and AM for the $8$ datasets from \citet{malioutov2013exact}. However, on the two larger datasets (MAGIC and Musk), RIPPER selects a rather large number of features, and further study on large datasets is needed to clarify the advantages and disadvantages of the algorithms. In addition, AM achieves the highest accuracy on Pima, and BCD obtains the highest accuracy on WDBC.

Below is an example of a learned rule that predicts Parkinson's disease from voice features. (It is consistent with known low frequency and volume change reduction in the voices of Parkinson's patients \cite{ramig2004parkinson}.) 

\begin{tabbing}
\=IF\quad\=1. voice fractal scaling exponent $>-6.7$; OR\\
  \>\>2. \= max vocal fundamental frequency $<236.4$ Hz; AND\\
  \>\>\> min vocal fundamental frequency $<181.0$ Hz; AND\\
  \>\>\> shimmer:DDA $<0.0361$; AND\\
  \>\>\> recurrence period density entropy $<0.480$;\\
  \>THEN this person has Parkinson's.
\end{tabbing}

\section{Conclusion}\label{sec:Conclusion}
Motivated by the need for interpretable classification models, this paper has provided an optimization-based formulation for two-level Boolean rule learning. These complement the more heuristic strategies in the literature on two-level Boolean rules. Numerical results show that two-level Boolean rules typically have considerably lower error rate than one-level rules and provide very good tradeoffs between accuracy and interpretability with improvements over state-of-the-art approaches.

\section*{Acknowledgment}
The authors thank V. S. Iyengar, A. Mojsilovi\'{c}, K. N. Ramamurthy, and E. van den Berg for conversations and support.

\bibliography{ReferenceList}
\bibliographystyle{icml2016}

\end{document}